\documentclass{article}

\usepackage[preprint]{neurips_2024}

\makeatletter
\renewcommand{\@noticestring}{}
\makeatother

\usepackage[utf8]{inputenc} 
\usepackage[T1]{fontenc}    
\usepackage{hyperref}       
\usepackage{url}            
\usepackage{booktabs}       
\usepackage{amsfonts}       
\usepackage{nicefrac}       
\usepackage{microtype}      
\usepackage{xcolor}  
\usepackage{graphicx}
\usepackage{amsmath}
\usepackage{subcaption}
\usepackage{mathbbol}
\usepackage[normalem]{ulem}

\newcommand{\nearbest}[1]{%
  \begingroup
  \setlength{\ULdepth}{2pt}%
  \renewcommand{\ULthickness}{0.35pt}%
  \uline{#1}%
  \endgroup
}

\title{Benchmarking Deep Learning Models for Raman Spectroscopy Across Open-Source Datasets}

\author{%
Adithya Sineesh$^{1} \thanks{Corresponding author.}$ \quad Akshita Ramya Kamsali$^{1}$\\
$^1$Purdue University, West Lafayette \\
\texttt{\{asineesh,akamsali\}@purdue.edu}
}

\begin{document}

\maketitle

\begin{abstract}
Deep learning classifiers for Raman spectroscopy are increasingly reported to outperform classical chemometric approaches. However, their evaluations are often conducted in isolation or compared against traditional machine learning methods or trivially adapted vision-based architectures that were not originally proposed for Raman spectroscopy. As a result, direct comparisons between existing deep learning models developed specifically for Raman spectral analysis on shared open-source datasets remain scarce. In this work, we focus on supervised Raman spectra classification where each spectrum is assigned to a predefined material, bacterial/yeast isolate, drug treatment or pharmaceutical compound. To the best of our knowledge, this study presents one of the first benchmarks comparing three or more published Raman-specific deep learning classifiers across multiple open-source Raman datasets. We evaluate five representative Deep Learning (DL) architectures along with two conventional Machine Learning (ML) methods under a unified training and hyperparameter tuning protocol across three open-source Raman datasets selected to support standard evaluation, fine-tuning, and explicit distribution-shift testing. In this comparative study, we primarily focus on classification because the selected open-source datasets provide classification annotations, while annotations for complete structure elucidation are not available. We report classification accuracies and macro-averaged F1 scores to provide a fair and reproducible comparison of the supervised ML and DL models for Raman spectra based classification.
\end{abstract}

\section{Introduction}\label{sec1}
Raman spectroscopy is a non-destructive characterization technique, where the spectra encode vibrational signatures of molecular bonds. This non-destructive capability enables applications spanning biomedical diagnostics \cite{lin2025multi}, pharmaceuticals \cite{roggo_identification_2010}, materials identification \cite{liu2017deep} and food quality assessment \cite{chai2024novel}. Despite this breadth, Raman spectra are a weak scattering phenomenon, which leads to captured spectra often containing artifacts such as distortion by noise, background fluorescence, cosmic ray impacts and other environmental factors. These artifacts can dominate the class distinguishing features in the signal which complicates statistical learning for material identification \cite{vulchi2024artifacts}. Another challenge for automated identification of materials is the spectral overlap of the characteristic peaks of different components in the substance. Beyond additive noise, Raman spectra vary across systems and their configurations leading to significant distributional shifts that undermine the performance guarantees established on the test datasets when the models are deployed in the real world \cite{guo2020comparability}. At the same time, constructing large, labeled Raman spectroscopy datasets is costly as hardware is expensive and careful sample preparation, calibration and repeated measurements are labor intensive. Furthermore, annotation and assigning peaks to specific bonds or chemical groups often requires domain expertise which is often not available at scale.
Machine Learning has therefore been increasingly adopted for material identification. Early works showcased the effectiveness of traditional chemometric approaches like Principal Component Analysis (PCA), Linear Discriminant Analysis (LDA) and Support Vector Machines (SVMs) in material classification using Raman spectra \cite{monavar2013determining} \cite{roggo_identification_2010}. More recent approaches leverage deep neural models, particularly one-dimensional convolutional networks \cite{liu2017deep} \cite{dong2019practical} \cite{kirchberger2020towards} and attention-based architectures \cite{liu2023classification} \cite{wang_deep_2024} \cite{koyun_ramanformer_2024} and often report substantial gains over classical baselines. However, most of the existing works do not compare their proposed methods against other published deep learning approaches for Raman spectra classification. Such comparisons are hindered by challenges in data reproducibility, dataset accessibility, and limited data-sharing frameworks \cite{coca2025artificial}.

Benchmarking under a standard evaluation framework allows for fair and meaningful comparisons across different architectures and algorithms. Such evaluations help researchers identify which strategies are effective across different datasets and problem settings. These established reference points can serve as an excellent starting point for advancing state-of-the-art performance. In this work, we benchmark five supervised deep learning models proposed for Raman spectra classification, along with two conventional machine learning models, under a controlled and reproducible training and evaluation protocol across three open-source Raman datasets and report their accuracy and macro-averaged F1 score. We focus on classification because it is a widely used and well defined Raman spectroscopy task for material identification. The selected open-source datasets provide labels that directly support this task. Other Raman analysis tasks such as mixture analysis and complete structure elucidation require different annotations, which are not available for all the datasets at the time of this study and are therefore beyond the scope of the present benchmark. In subsequent sections, we outline the motivation behind the selection along with the details of the deep learning models and the datasets.

\section{Literature Survey}

Machine learning techniques have been widely used for multi-class classification, where each spectrum is assigned to exactly one material. It has also been applied to the broader problem of multi-label classification, where the constraint of a sample containing only one target is relaxed. This task involves associating each Raman spectrum with multiple materials simultaneously. Multi-label classification with material concentration estimation further extends this task by estimating the mixing ratios in addition to predicting the components present in the sample. In the following subsections, we review the prior work in all these different types of Raman spectra classification. We organize the literature from foundational statistical approaches to modern deep learning based models.

\subsection{Classic Chemometrics}
Early Raman-based classification pipelines used linear dimensionality reduction to compress the spectrum and then trained shallow discriminative models with limited number of samples. For example, Rebro{\v{s}}ov{\'a} et al. categorized staphylococci into 16 strains using Principal Component Analysis (PCA) for feature compression followed by 1-NN classification on 277 spectra with 5-fold cross-validation \cite{rebrovsova2017rapid}. Similarly, Monavar et al. leveraged PCA-derived representations with shallow Artificial Neural Networks (ANNs) and Linear Discriminant Analysis (LDA) for caviar type identification despite having only 93 spectra \cite{monavar2013determining}. In pharmaceuticals, Roggo et al. used Support Vector Machine (SVM) based hierarchies to first identify tablet product families and then refine predictions to formulation categories via correlation-based methods and additional SVMs, using 25 product families comprising of 44 formulations (15 spectra per formulation) \cite{roggo_identification_2010} in total. Generally, these classic chemometrics studies used compressed feature spaces coupled with lightweight classifiers. Recent years have seen a gradual shift from such traditional pipelines toward modern deep learning approaches that aim to learn representations directly from raw (or minimally processed) spectra and better tolerate perturbations expected in Raman spectra.

\subsection{Deep Learning Models Evaluated on Small Experimental Raman Datasets}
Deep Learning approaches for Raman spectroscopy mainly involve developing 1-D Convolutional Neural Networks (CNNs) or transformers and comparing them against classical baselines.  An early example is the Deep CNN mineral classifier by Liu et al. \cite{liu2017deep}, which was evaluated against KNN/SVM/Random Forest on RRUFF-derived datasets \cite{lafuente20151}. While such studies established the potential of CNN-based models for Raman analysis, their evaluations were often conducted on test sets comprising of a limited number of experimentally acquired spectra. For example, Dong et al. designed a CNN with constrained kernels that emulate standard denoising and baseline-correction operations. The model achieved superior binary classification of human vs. animal blood (109 test spectra spanning humans, dogs, and rabbits using a 67:33 train-test split), outperforming SVM and PLS-DA \cite{dong2019practical}. Similarly, Kirchberger-Tolstik et al. developed a CNN model to predict the severity of Ulcerative Colitis using the Raman spectra of the colon biopsy \cite{kirchberger2020towards}. The model was assessed using patient level cross-validation on 227 Raman spectra acquired from 42 patients and not compared with any other approaches. 

A Locally Connected Neural Network (LCNN) designed by Houston et al. \cite{houston2020robust} was used to detect chlorinated solvents using the Raman spectra of the samples. Despite outperforming SVM, KNN, Decision Tree, Gaussian Naive Bayes and Fully Connected Neural Network (FCNN), the evaluation was performed on a small real dataset of 58 test spectra.  In another example, the classification of three marine pathogen strains was achieved with the help of Generative Adversarial Networks \cite{yu2020classification} by Yu et al. The independent Generator-Discriminator for each pathogen strain were trained on the corresponding 50 experimental spectra. Testing was conducted on 60 real spectra, achieving a reported 100\% classification accuracy, though the method was not compared against competing approaches. 

RaMixNet I and II \cite{mozaffari_convolutional_2021} were developed for multi-label material identification and multi-label classification with concentration estimation respectively. These models were trained primarily on synthetic spectra generated by linear combinations of four pure compounds with additional baseline drift and spectral augmentations. Evaluation on real data was restricted to only six experimentally measured mixtures where they outperformed correlation-based methods and Partial Least Squares Regression.

While these works showed the promising performance of deep learning methods, their heavy reliance on small-scaled experimental datasets raises questions about their efficacy during real-world deployment.

\subsection{Synthetic Data Driven Evaluation of Raman Classification Models}

Subsequent works tried to alleviate the limitation of small-scaled experimental datasets by making extensive use of synthetic data. For example, Qi et al. developed a CNN model to classify 10 different 2-D materials \cite{qi2024deep} like Graphene using limited number of real Raman spectra along with large quantities of synthetic data generated by Denoising Diffusion Probabilistic Models (DDPMs) \cite{ho2020denoising}. The dataset consisted of 10,000 synthetic spectra and 594 experimental spectra. The authors compared their model to SVM, ANN, KNN methods using a 10 fold cross-validation scheme. Similarly, Hamed Mozaffari and Tay trained a model consisting of a single 1-D convolutional layer with two linear layers \cite{hamed_mozaffari_overfitting_2022} on 40,000 synthetic spectra generated by augmenting 5 real Raman spectra of 5 different materials. This model outperformed several other published Raman ML models on this 5-category classification problem. However, the test dataset comprised of 5000 synthetic spectra generated by augmenting the same 5 gold-standard spectra as the train dataset. 

DeepCID \cite{fan_deep_2019} further exemplifies this evaluation protocol. The framework is comprised of a suite of CNN models that each predict if a specific material is present in the sample or not. The models are trained on synthetically generated spectra that were generated by applying augmentations to 167 real spectra of pure common pharmaceutical raw materials. It outperforms traditional chemometric methods and a FCNN on synthetic data and limited real Raman spectra of mixtures. Overall, synthetic data is often used as a remedy for the scarcity of experimentally acquired Raman spectra. But evaluating models on synthetic test datasets leads to the same limitations of using small real test datasets i.e. uncertainty about their performance on unseen samples during real-world deployment.

\subsection{Evaluation Limited to Classical or Generic Deep Learning Baselines}

In addition to the limitations due to the use of small experimental or synthetic test datasets, several deep learning models for Raman spectroscopy have been proposed without systematic comparison to existing Raman-specific deep learning approaches. Evaluations are often restricted to classical machine-learning baselines or generic image-based architectures which makes it difficult to assess the merits of the proposed methods within the broader Raman spectroscopy literature. For example, Maruthamuthu et al. used a ResNet-18 \cite{he2016deep} inspired CNN model for the detection of microbial contamination \cite{maruthamuthu2020raman}. The dataset contained 6000 real Raman spectra each of Chinese Hamster Ovary (CHO) cells, which are widely used in the pharmaceutical industry, along with 12 microbes that are the common contaminants of CHO and 3 mixtures of CHO and contaminant microbes. The model was trained using a 5-fold cross-validation scheme but not compared against any other approaches, and the exhaustive dataset is not public. 

Similarly, Primrose et al. adapted the VGG13 \cite{simonyan2015deepconvolutionalnetworkslargescale} architecture for Raman spectra classification by replacing all the 2-D convolutional layers with 1-D convolutional layers \cite{primrose_one_2022}. This model was then trained on a synthetically mixed dataset and was evaluated on limited data collected by the First Defender Raman spectrometer and showed >90\% detection rates for certain explosive materials and their precursors, outperforming the algorithm used by the spectrometer. This work was extended \cite{primrose_one-dimensional_2023} to include additional non-explosive materials and the evaluation was performed over 10,000 real Raman spectra but the dataset was not made public and the model was not compared against other published Raman-specific approaches.

A similar evaluation pattern is observed for Raman Spectral Translation (RST) \cite{wang_deep_2024}, which was developed for multi-label classification by combining the ideas from CNNs and transformers.  However, the comparisons were limited to a generic CNN and DenseNet \cite{huang2017densely}, rather than against established Raman-domain architectures, on a real test dataset. Likewise, a transformer model for classification of deep-sea cold seep bacteria \cite{liu2023classification} and a CNN model for classification of plastics \cite{qin2024deep} were compared using experimental test spectra against 1-D variants of AlexNet \cite{krizhevsky2012imagenet}, ResNet and SVM, LDA and Decision Trees respectively.

This trend extends to biomedical and agricultural applications as well. The RFBC \cite{chai2024novel} model used a hybrid CNN-LSTM architecture incorporating Fourier-domain features and outperformed PCA-SVM, PCA-KNN and PCA-XGBoost in detecting different brands of rice on a test dataset containing experimental Raman spectra. Similarly, Y. Lin et al. developed a ResNet-18 based model for cancer detection \cite{lin2025multi} using 2-D transforms of 1-D serum Raman spectra (CWT/heatmaps). The authors evaluated the model only against generic image classifiers like AlexNet, VGG16, DenseNet. Kok et al. designed another ResNet-based model, with multi-channel inputs spanning raw spectra plus multiple pre-processed views, \cite{kok2024classification} for osteoarthritis cartilage classification and compared the model against baseline CNNs. Similarly, Ullah et al. developed a Multi-Layer Perceptron (MLP) to detect tuberculosis from blood serum samples \cite{ullah_evaluating_2022} but did not compare the model with any other approaches. 

Du et al. developed another shallow CNN to classify Bacillus spores \cite{du_accurate_2023} but only compared the model to traditional ML approaches like SVM. RaT and RaST were transformer and Swin-Transformer \cite{liu2021swin} based architectures proposed for classifying lactic acid bacteria into 14 different strains \cite{wang2026deep}. These models were evaluated on an experimental test dataset against an adapted ResNet model, SVM, LDA, KNN and XGBoost.

Likewise, the evaluation is restricted to MLP, Least squares, modified VGG11 and ResNet-50 for the RamanFormer \cite{koyun_ramanformer_2024}, which was proposed to quantify the presence of Methanol, Isopropyl Alcohol and Ethanolamine in the Raman spectra of the mixture. Another example is ConInceDeep \cite{zhao_conincedeep_2023}, a multi-label classification model that combines Continuous Wavelet Transform (CWT) representations with convolution-based Inception Modules \cite{szegedy2016rethinking}. It was evaluated against ablated variants of its own proposed architecture.

Collectively, these works demonstrate a huge variety of deep learning approaches developed for Raman spectroscopy across different domains. However, the lack of benchmarking against established Raman-specific models hampers the ability to draw meaningful conclusions across multiple papers. 

\subsection{Limited Benchmarking}
Only a few works in Raman deep learning literature attempt explicit benchmarking against previously published models. However, the scope of the comparison is often limited in these cases.

RamanNet was proposed as a general Raman spectra classifier \cite{ibtehaz2023ramannet} and was evaluated on open-source datasets like the COVID-19 Raman dataset \cite{yin2021efficient}, Melanoma dataset \cite{erzina2020precise}, RRUFF Mineral database \cite{lafuente20151} and the Bacteria-ID dataset \cite{ho2019rapid}. While RamanNet outperformed the baseline for each dataset, the benchmarking was limited as the comparison was just against one model per dataset. A Scale-Adaptive deep neural network (SANet) was designed for identifying the isolate and the empiric treatment for the sample based on its Raman spectrum \cite{deng2021scale}. The model was trained and evaluated on the Bacteria-ID dataset and compared against the model presented in that work \cite{ho2019rapid} and traditional ML methods like SVM and Linear Regression rather than a broader set of Raman-specific deep learning models. 

The Wavelet Packet transform and Gramian Angular field (WPGA) algorithm was developed by Liu et al. to generate 2-D spectrograms from 1-D Raman spectral data \cite{liu2024rapid}. The authors then showed that a ResNet based model using these 2-D spectrograms outperformed several published Raman classifier models \cite{ho2019rapid} \cite{ibtehaz2023ramannet} \cite{zhou2022ramannet} on the open-source Bacteria-ID dataset \cite{ho2019rapid}. However, the official implementation of the model is not publicly available, and the paper and its supplementary material do not provide sufficient architectural details to allow for an exact reproduction of the proposed network.

Lange et al. published an open-source dataset containing 6960 Raman spectra of mixtures containing 8 different metabolites in varying concentrations \cite{lange2025deep}. The authors compared 11 different models on it but only one of those models was published for Raman spectra classification.

Overall, the prior work in Raman spectra classification shows tremendous architectural innovation but suffers from limitations due to: (i) small experimental datasets or synthetically generated test sets, (ii) comparisons restricted to classical machine-learning baselines or trivially adapted vision models. This complicates meaningful cross-paper performance assessment. In this work, we present reproducible benchmarking under consistent experimental settings for five supervised Raman-specific DL models and two conventional ML models across three open-source Raman spectral datasets.  

\section{Benchmark models}
In the previous section, we highlighted the numerous approaches employed over the years for material classification and material analysis of samples based on their Raman spectrum. In this work, the benchmarking is limited to just supervised deep learning models for multi-class classification. We chose five models, detailed in subsequent sections, which represent a variety of architectural designs, complexities and sizes as shown in Table \ref{tab:model_info}. All these models were trained using Cross-Entropy Loss, unless mentioned otherwise. We also employ Random Forest and Support Vector Classifier (SVC) to investigate any performance differences between these conventional ML models and the five Raman-specific DL models.

\begin{table}[!ht]
\caption{The five chosen models in terms of parameter count and multiply-accumulate (MAC) operations for a Raman spectrum of length 1024 with 15 output classes \label{tab:model_info}}
\begin{center}
\begin{tabular*}{0.7\textwidth}{@{\extracolsep\fill}lcc@{\extracolsep\fill}}%
\hline
\textbf{Model} & \textbf{Number of parameters (M)} & \textbf{MACs (M)}\\ 
\hline
Deep CNN \cite{liu2017deep} &  15.91 & 21.77 \\ 
SANet \cite{deng2021scale} &  2.23 & 102.48 \\
RamanNet \cite{ibtehaz2023ramannet} & 0.72 & 0.72  \\
Transformer \cite{liu2023classification} & 85.17 & 769.88 \\
RamanFormer \cite{koyun_ramanformer_2024} & 4.31 & 24.33 \\ 
\hline
\end{tabular*}
\end{center}
\end{table}

\subsection{CNN-based models}
The Deep CNN model \cite{liu2017deep} was one of the first Deep Learning models introduced for Raman spectra classification and is based on the famous LeNet-5 Architecture \cite{lecun2002gradient}. It consists of three 1-D convolutional layers with kernel sizes of 21, 11 and 5 interleaved with pooling layers that reduce the spatial dimension of the spectra. The features generated by these layers are then passed through a dense linear layer followed by a classification head, which contains the same number of output nodes as the number of classes in the dataset.

Convolution has the property of shift equivariance, which is desirable for images but detrimental for Raman spectra where the position of the peak plays a vital role in identifying the material. The RamanNet \cite{ibtehaz2023ramannet} model was developed to be free from this translational equivariance. The spectrum is split into overlapping sliding windows of a fixed size and each window is passed through its own multi-layer perceptron. This operation is similar to convolution but with different kernels applied to different spatial locations. The outputs for all the windows are then concatenated and passed through another multi-layer perceptron and a linear embedding layer to obtain the feature representation of the spectra. These features are then fed into a classification head to obtain the predicted label. RamanNet is trained using a linear combination of triplet loss on the feature representation of the spectra and Cross-Entropy loss on the predicted class labels.

At each convolutional layer in a CNN, the size of the Receptive Field (RF) is constant. This property was said to be undesirable for capturing the peaks of a Raman spectrum, which are of different widths, and thus the Scale Adaptive Network (SANet) \cite{deng2021scale} was developed. It consists of five Multi-Scale Blocks for feature extraction with each Block consisting of six 1-D convolutional layers having increasing kernel sizes from 3 to 13. This allows for each Block to capture the features at different scales of RF. These features are stacked along the channel dimension followed by channel-attention and point-wise convolution to extract only the relevant features while reducing the channel dimension. The features generated by the sequence of MultiScale Blocks are then flattened and passed through the classification head.

\subsection{Transformer-based models}

Recent years have seen the proliferation of attention-based mechanisms for vision and language tasks. The transformer\cite{vaswani_attention_2017} architecture, which is solely based on attention, has also been adopted for the classification of Raman spectra. This development was only natural as the attention mechanism of the transformer was originally designed to model sequential data. The earliest such attempt involved simply adapting the Vision Transformer (ViT)\cite{dosovitskiy2020image} for the 1-D spectral domain \cite{liu2023classification}. The spectrum is split into patches and then passed through a linear layer to map them to tokens of dimension 768. A learnable class token of the same dimension is appended to this sequence of tokens followed by the addition of Position encoding to track the sequence of tokens. The model consists of 12 transformer blocks, each comprising of a 12-head self-attention layer and a multi-layer perceptron. The former captures the relationships between the different tokens in the sequence and the latter helps to generate higher-level features from each token. At the output of the transformer blocks, the embedding of the class token serves as the aggregate representation of the input sequence. Therefore, this embedding is passed through the classification head. This model shall be referred to as the Transformer in the subsequent sections.

The RamanFormer \cite{koyun_ramanformer_2024} further modifies the transformer architecture for processing Raman spectra. The generation of tokens from the spectrum remains the same, only without the class token and with a reduced embedding dimension of 256. This sequence of 8 tokens are passed through just three transformer blocks, each containing a 8-head self-attention layer and a MLP. The output token sequence is fed through two strided convolutional layers to capture the spatial hierarchy. The features are then pooled along the temporal dimension followed by passing them through a dense layer and classification head. Although RamanFormer was originally proposed for Mixture analysis of Raman spectra, we use it for classification tasks. The underlying architecture remains the same with Cross-Entropy loss being used instead of the original L1 loss.

\section{Datasets} \label{sec4}
\begin{figure*}[ht]
\centering
\includegraphics[width=\textwidth]{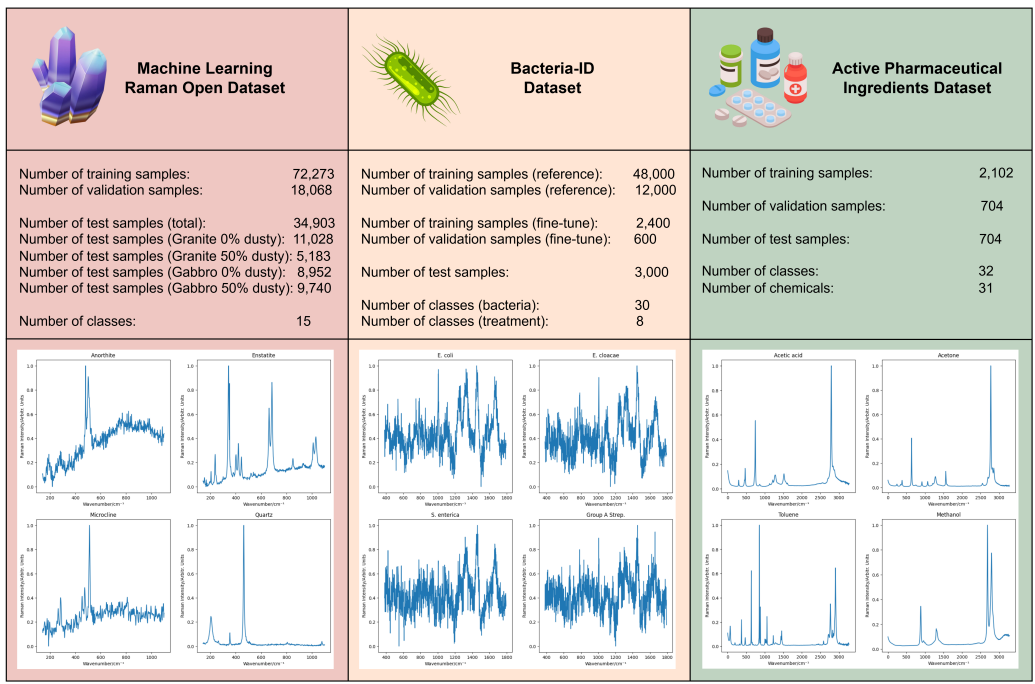}
\caption{\textbf{Dataset overview and qualitative comparison across domains.} Top row: the three Raman spectroscopy datasets used in this study: 1. \textbf{MLROD} (mineral spectra), 2. \textbf{Bacteria-ID} (spectra of bacterial species/strains), and 3. \textbf{API} dataset (spectra of active pharmaceutical ingredients). Middle row: dataset-level statistics summarizing scale and label structure including total number of classes. Bottom row: \textbf{representative spectra} randomly sampled from each dataset (intensity vs. Raman shift ($\text{cm}^{-1}$)).
\label{fig3}}
\end{figure*}

We surveyed publicly available Raman spectroscopy datasets and selected three that span distinct application domains and exhibit varying degrees of distribution shift between the training and test sets. Specifically, we choose the MLROD (minerals) \cite{berlanga_convolutional_2022}, Bacteria-ID \cite{ho2019rapid} (biomedical), and API \cite{flanagan2025open} (pharmaceutical) datasets as shown in Figure \ref{fig3}. The MLROD test set was generated under different conditions than the training set, enabling evaluation under distribution shift. The Bacteria-ID dataset consists of a reference set suitable for pretraining and a smaller fine-tuning set that contains the same degradation in optical system efficiency as the test dataset. Meanwhile the test split of the API dataset is from the same distribution as the train and validation split. However, unlike the previously mentioned datasets, the API dataset does not provide predefined train and test splits and these splits must be constructed by the user. We now discuss the three datasets and their corresponding acquisition method and the criteria we test for. 

\subsection{Machine Learning Raman Open Dataset (MLROD)} 
\begin{figure*}[ht!]
\centering
\includegraphics[width=\textwidth]{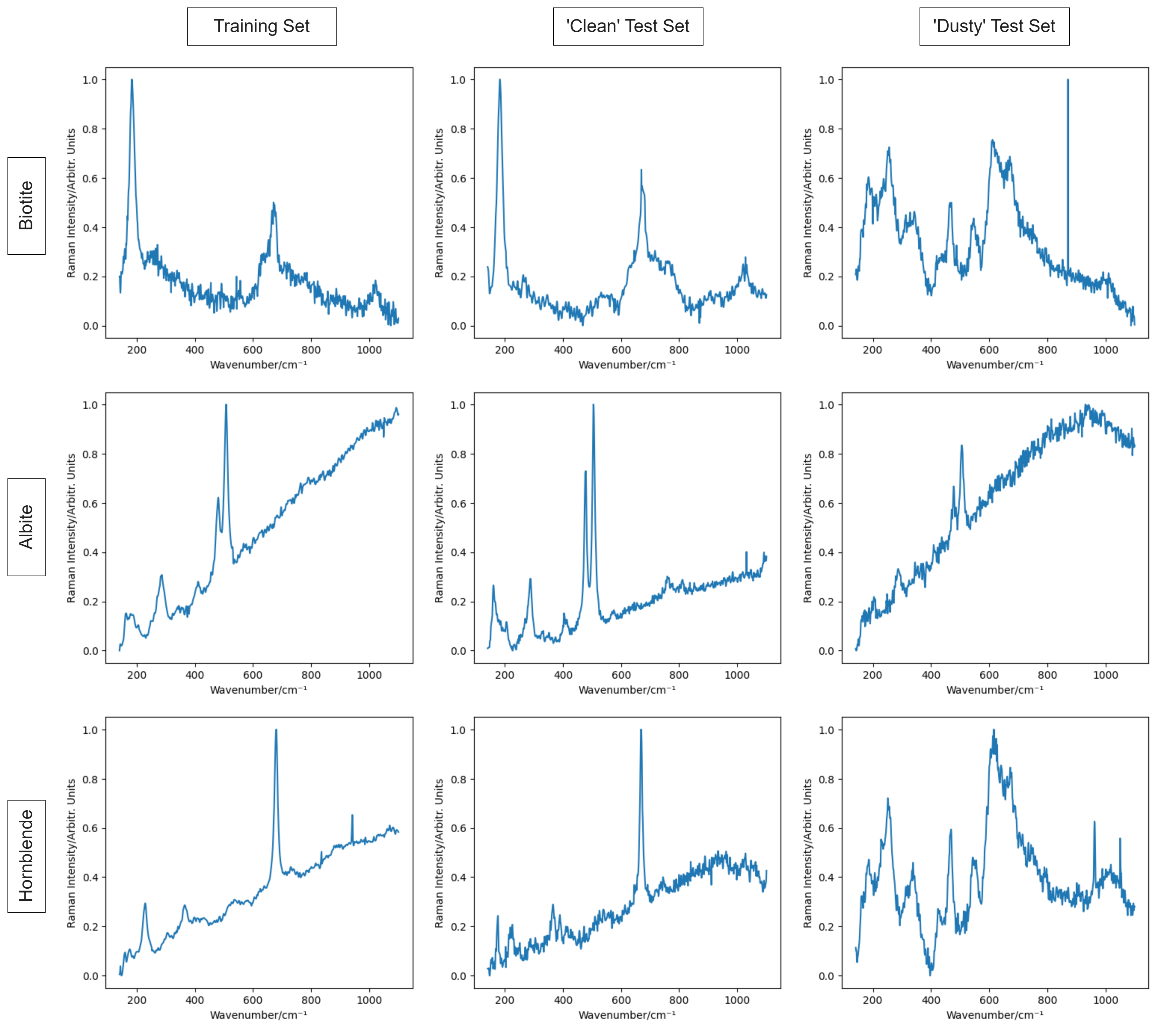}
\caption{\textbf{MLROD spectral shift across evaluation conditions (“clean” vs. “dusty”).} 
Representative Raman spectra for three minerals (\textbf{Biotite, Albite, Hornblende}) shown across the \textbf{training set} (left), \textbf{`clean' test set} (middle), and \textbf{`dusty test' set} (right). Each row corresponds to the same material label across splits; each trace is a single example spectrum plotted as Raman intensity (arb. units) versus wavenumber ($cm^{-1}$). While the training and `clean' test spectra largely preserve characteristic peak locations and relative band structure, the `dusty' split exhibits pronounced contamination artifacts such as elevated and drifting baselines, broadened features, and spurious high-intensity components. This illustrates a significant distribution shift that challenges model generalization.
\label{fig1}}
\end{figure*}

MLROD is a high-volume Raman dataset created for material detection on Mars. It contains 89,121 labeled spectra spanning 12 pure mineral classes and 3 binary 1:1 powder mixtures. Mineral classes are: Quartz, Albite, Anorthite, Microcline, Hornblende, Biotite, Muscovite, Forsterite, Augite, Enstatite, Calcite, Gypsum; mixtures are Quartz+Albite, Forsterite+Augite, Forsterite+Albite. It also consists of a separate set of test spectra (rocks, with and without dust): 39,720 spectra from Gabbro and Granite slabs, measured under clean (0\% dust) and dusty (~50\% Basaltic dust coverage) conditions (e.g., Gabbro: 8,952 clean / 9,740 dusty; Granite: 11,028 and 10,000 across the two conditions as listed) \footnote{When we downloaded the data from the official website, the training dataset had 90,341 spectra. Several of the spectra in the Granite 50\% dust test dataset have a label that does not correspond to any of the labels in the training dataset. We have ignored these samples for the purposes of evaluation and therefore the size of the Granite 50\% dust test dataset is 5,183 and the size of the overall test dataset is 34,903.}. The ``dusty'' regime was created using basaltic dust to mimic the obscuration in Martian conditions. Figure \ref{fig1} shows the difference in the Raman spectra collected in the ``train'', ``clean'' test and ``dusty'' test regimes. The dataset was collected on Horiba LabRAM HR Evolution single stage spectrometer with 532 nm excitation with no pre-processing steps and only standardizing the axis via interpolation and trimming. 

\subsection{Bacteria-ID Dataset} It is a bacterial (mostly monolayer and single-cell) Raman spectroscopy dataset intended for pathogen identification and downstream grouping by 8 antibiotic treatments. The dataset provides a reference dataset across 30 bacterial plus yeast isolates with 2000 spectra per isolate for training. Additionally, it contains fine-tuning and test sets each with 100 spectra per isolate. The dataset was collected on Horiba LabRAM HR Evolution Raman microscope at 633nm excitation with polynomial background correction and per-spectrum min-max normalization to [0,1] range.

\subsection{Active Pharmaceutical Ingredients (API) Dataset}
The API dataset is an open-source dataset with 3510 spectra spanning 32 pure compounds measured on Raman Rxn2 analyzer at 785nm excitation. The dataset provided comes with automatic instrument pre-treatment which includes dark noise subtraction, cosmic ray filtering and intensity correction with no other pre-processing steps.

\section{Methodology} \label{sec5}


In this section, we outline the training methodology and evaluation approach used to benchmark the five Raman-specific deep learning models, along with two conventional machine learning methods, on the three open-source Raman spectroscopy datasets. For consistency, we use the same optimization procedure (Adam), model-selection criterion and stopping rules across all deep learning models, while tuning hyperparameters independently for each model-dataset task pair using an identical search procedure. We also include Random Forest and a Support Vector Classifier (SVC) as conventional machine learning methods to evaluate how their performance compares with that of the five Raman-specific deep learning models.

\begin{table}[!ht]
\caption{List of hyperparameters evaluated for the grid search, applied independently to each of the five DL models on all three datasets, with separate tuning for every model–dataset pair to ensure fair comparisons \label{tab:hyper}}
\begin{center}
\begin{tabular*}{0.48\textwidth}{@{\extracolsep\fill}lc@{\extracolsep\fill}}%
\hline
\textbf{Hyperparameters} & \textbf{Values} \\ 
\hline
Batch Size  &  32, 128, 512 \\ 
Learning Rate &  1e-3, 1e-4, 1e-5 \\
\hline
\end{tabular*}
\end{center}
\end{table}

\subsection{Preprocessing} For all datasets, we applied only intensity scaling to the raw spectra to improve the numerical stability of the conventional ML and DL models. We intentionally skip any further pre-processing steps such as baseline correction, fluorescence removal, scatter correction, or denoising. Although several procedures exist for each of the processing steps, it is difficult to engineer an effective pre-processing pipeline that can be applied to all the datasets. This is because it is challenging to identify which combination of methods is optimal, as in most cases they lead to worse model performance \cite{engel2013breaking}. Deep learning based approaches for artifact removal have also been developed recently \cite{gebrekidan2021refinement} \cite{sjoberg2025radar} but they add to the time complexity of the pre-processing pipeline and can obscure whether performance gains arise from the classifier architecture or from learnable pre-processing. Our goal is to benchmark model behavior under a minimal, reproducible pre-processing regime.

\subsection{Hyperparameter tuning and Final evaluation runs}
For each deep learning model-dataset task pair, we performed hyperparameter tuning via $3 \times 3$ grid search over the ranges listed in Table \ref{tab:hyper}. For the Random Forest and the SVC, we performed hyperparameter tuning via  grid search over the ranges listed in Supplementary Table 1 and Supplementary Table 2.  We select the optimal hyperparameters based on the model with the best validation performance. This tuning was performed independently per dataset task to avoid transferring dataset task-specific choices across domains. After hyperparameter tuning, each model-dataset task pair was retrained for five independent final evaluation runs using its corresponding optimal hyperparameters.

We used the same maximum epoch budget and early-stopping patience across all DL models to maintain a unified benchmark protocol for hyperparameter tuning and final evaluation runs. This choice is intended to compare the published architectures under identical constraints, rather than to independently optimize the training configuration for each DL model. We acknowledge that this shared protocol may not be optimal for all architectures, particularly the transformer-based models, whose performance may depend on learning-rate warmup, weight decay, dropout tuning, and longer training schedules.

\subsection{MLROD}
We divided the MLROD training dataset into the train and validation datasets using a random 80:20 train-val split. For the DL models, the training was for up to 40 epochs, with early stopping applied if the validation accuracy didn't improve for 10 consecutive epochs. We then evaluated the DL model with the best validation accuracy on the hold-out test dataset. The Random Forest and SVC were fit to the train split and evaluated on the hold-out test set. We also reported the accuracy separately for the clean Gabbro, clean Granite, dusty Gabbro and dusty Granite subsets of the test dataset. 

\subsection{Bacteria-ID}
The 3 relevant subsets used from the Bacteria-ID dataset to benchmark the selected models were the reference, fine-tune and test sets. We divided the reference and fine-tune sets into their corresponding train and validation datasets using a random 80:20 train-val split. 

For the DL models, the pretraining was for up to 40 epochs on the train split of the reference set, with early stopping applied if the accuracy did not improve on the validation split of the reference set for 10 consecutive epochs. The DL model with the best reference validation accuracy was then trained on the fine-tune train split with a tenth of the learning rate used in pretraining. The fine-tuning was for up to 40 epochs with early stopping applied if the accuracy did not improve on the validation split of the fine-tune set for 10 consecutive epochs. We then reported the accuracy on the test set using the model with the best fine-tune validation accuracy.

Since Random Forest and SVC do not have a fine-tuning stage, we introduced an additional hyperparameter indicating whether the training set consisted of only the reference set, only the fine-tune set or both. For final evaluation, the conventional ML models were fit on the optimal train set and evaluated on the test set.

All the models were trained separately on the two tasks of classifying the isolate and classifying the empiric treatment.

\subsection{API Dataset} For the API dataset, we generated the test split by randomly sampling 22 disjoint samples each for all the 32 classes and assigned the remaining samples to the train/val set. This approach approximately corresponds to a 80:20 train/val-test split. The train and validation sets were generated by using a subsequent random 80:20 split. 

For the DL models, the training was capped at 40 epochs with early stopping if the validation accuracy of the DL model did not improve in the past 10 epochs. We then evaluated the DL model with the best validation accuracy on the test split. The conventional ML models were fit on the train split and evaluated on the test split.

\begin{figure}[ht!]
\centerline{\includegraphics[width=\textwidth]{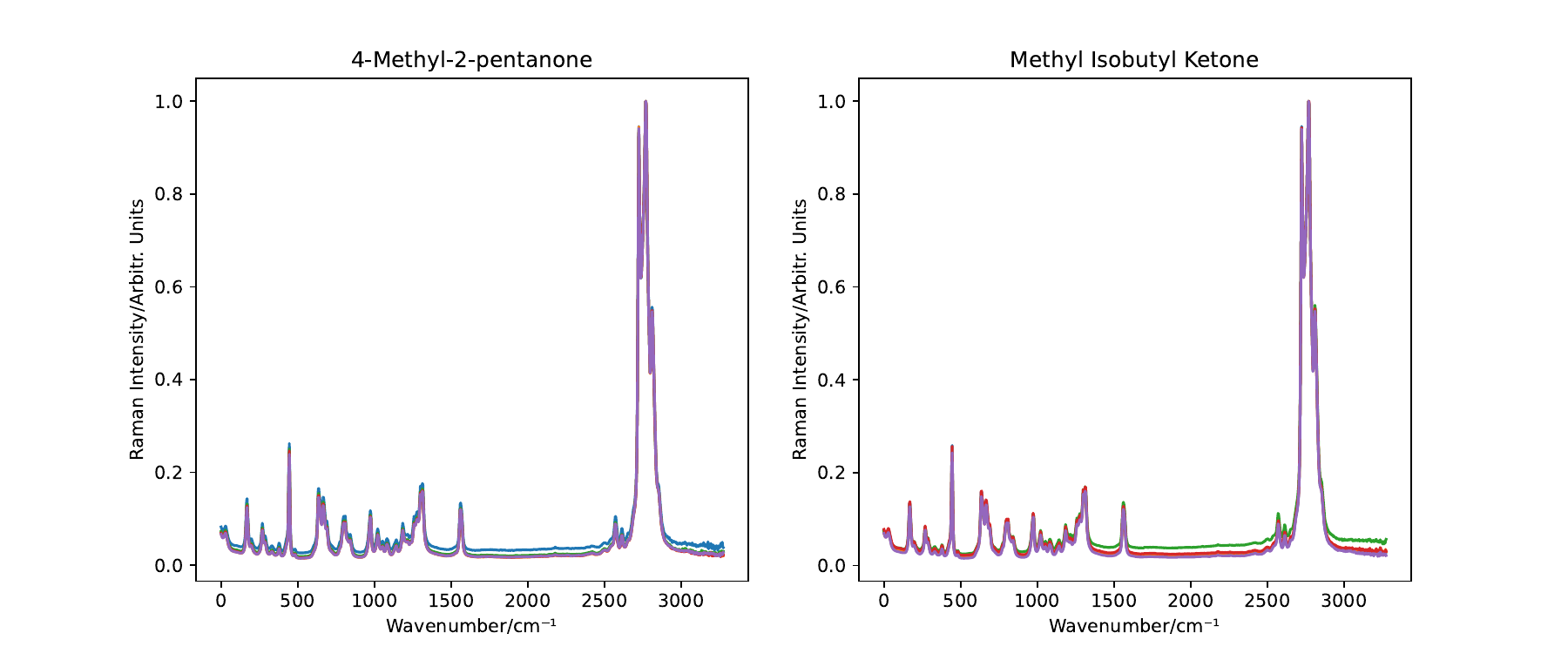}}
\caption{\textbf{API dataset label synonymy.} Overlaid Raman spectra (five randomly selected measurements per panel) for \textbf{4-methyl-2-pentanone} (left) and \textbf{methyl isobutyl ketone} (right), plotted as Raman intensity (arb. units) versus wavenumber ($cm^{-1}$). The near-identical signatures across the two panels reflect that methyl isobutyl ketone and 4-methyl-2-pentanone are chemically identical, illustrating potential label aliasing in the dataset and motivating label harmonization during our evaluation \label{fig2}}
\end{figure}

Although the dataset contains 32 labeled compounds, two of them, 4-methyl-2-pentanone and methyl isobutyl ketone, are chemically identical \footnote{https://pubchem.ncbi.nlm.nih.gov/compound/Methyl-Isobutyl-Ketone}. This is also highlighted in the dataset, where the samples of both these compounds have highly similar Raman spectra, as shown in Figure \ref{fig2}. Therefore, we report the results under two evaluation protocols. First, a 32 class setting that follows the dataset labels. Second, a 31 class setting where these two labels are merged post-hoc during evaluation. All the models are trained under the 32 class problem.

\subsection{Implementation Details} All experiments were run on a server with an Intel Xeon CPU (3.0~GHz), 64~GB RAM, and an NVIDIA RTX A5000 GPU. SANet uses the authors’ official PyTorch implementation, while the other DL models were reimplemented based on the details in the original papers or accompanying TensorFlow code. The conventional machine learning models were implemented using the scikit-learn Python library.

\section{Results} \label{sec 6}
For all the five Raman-specific DL models and two conventional ML models, we report their performance across the three datasets using accuracy and macro-averaged F1 score in the following subsections. Accuracy provides a simple and intuitive measure of overall correctness across all the samples and it is defined as follows:

\begin{equation}
    \text{Accuracy} = \frac{1}{N} \sum_{i=1}^{N} \mathbf{1} (\hat{y_i}=y_i )
\end{equation}

where $N$ is the total number of samples, $\mathbf{1}()$ denotes the indicator function, $y_i$ denotes the ground-truth class label and $\hat{y_i}$ denotes the predicted class label for the $i^{th}$ sample.

However, accuracy can be misleading in the presence of class imbalance. For example, strong performance on more frequent classes can mask poor performance on less frequent classes. To provide a more comprehensive and class-balanced assessment of model performance, we additionally report the macro-averaged F1 score.

The macro-averaged F1 score is the unweighted average of the F1 scores per class and is defined as follows:
\begin{equation}
    \text{macro F1}= \frac{1}{C}\sum_{c=1}^{C}\text{F1}_c
\end{equation}

The F1 score per class is defined as:

\begin{equation}
    \text{F1}_c=\frac{2*\text{TP}_c}{2*\text{TP}_c+\text{FP}_c+\text{FN}_c}
\end{equation}

where TP$_c$ is the number of true positives, FN$_c$ is the number of false negatives, and FP$_c$ is the number of false positives for class $c$.

By reporting both accuracy and macro-averaged F1 score, we capture the overall classification performance while ensuring that performance on minority classes is not obscured by the majority classes. This benchmark study evaluates single-label multiclass classification performance where the key question is whether the model assigns the correct class label. Accuracy and macro-F1 directly capture this, while AUC-ROC primarily evaluates the ranking of class scores across decision thresholds rather than the final predicted class assignments. Therefore we do not report AUC-ROC.

To provide simple baseline comparisons, we also report the macro-averaged F1 score and accuracy obtained on the test sets using a random classifier and a majority class classifier. The random classifier assigns labels to test samples by sampling uniformly from the class labels present in the training set. The majority class classifier assigns every test sample to the class that occurs most frequently in the training set. The random classifier was also evaluated over five independent random draws, while the deterministic majority class classifier has zero variance across runs.

\subsection{MLROD}
\begin{table*}[!ht]%
\caption{Test accuracy of the chosen models on different subsets of the MLROD test dataset. Values are reported as mean $\pm$ standard deviation across five independent final evaluation runs. $\uparrow$ indicates that higher values correspond to better performance for the reported metrics. Boldface highlights the best-performing model in each column. Underlined values indicate models whose mean plus one standard deviation equals or exceeds the highest mean in each column \label{tab:MLROD_results_acc}}
\resizebox{\textwidth}{!}{%
\begin{tabular}{llllll}
\hline
\textbf{Model} & \textbf{Granite 0\% dusty} $\uparrow$ & \textbf{Granite 50\% dusty} $\uparrow$ & \textbf{Gabbro 0\% dusty} $\uparrow$ & \textbf{Gabbro 50\% dusty} $\uparrow$ &  \textbf{Overall} $\uparrow$  \\
\hline
Random Classifier & 6.8\% (± 0.36) & 7.04\% (± 0.34) &  6.7\% (± 0.26) & 6.53\% (± 0.31)  & 6.66\% (± 0.19) \\
Majority Class Classifier & 53.96\% (± 0.0) & 38.01\% (± 0.0) & 0.0\% (± 0.0) & 0.0\% (± 0.0)  & 22.69\% (± 0.0)  \\
Random Forest & 92.16\% (± 0.19) & 66.84\% (± 0.62)  & 72.21\% (± 2.81)  & 41.82\% (± 0.42) & 69.23\% (± 0.78) \\
SVC & 91.25\% (± 0.21) & 61.5\% (± 0.27)  & 84.92\% (± 3.34)  & 39.76\% (± 0.24) & 70.84\% (± 0.93) \\
Deep CNN \cite{liu2017deep} & \textbf{97.66\% (± 2.18)} & 80.76\% (± 1.83) & \textbf{97.77\% (± 2.93)}  & 42.26\% (± 1.71)  & \nearbest{79.72\% (± 1.15)} \\
SANet \cite{deng2021scale} & 92.3\% (± 1.19) & \textbf{84.92\% (± 2.21)} & \nearbest{97.21\% (± 1.79)} & \textbf{48.87\% (± 2.4)}  & \textbf{80.34\% (± 0.88)}  \\ 
RamanNet \cite{ibtehaz2023ramannet} & 94.69\% (± 2.5)  & 80.22\% (± 3.22) & \nearbest{96.37\% (± 2.64)}  & 42.12\% (± 1.21) & 78.3\% (± 1.32) \\ 
Transformer \cite{liu2023classification} & 95.24\% (± 1.21) & 65.7\% (± 1.78)  & 81.82\% (± 6.41)  & 42.61\% (± 3.73)  & 72.72\% (± 2.41) \\ 
RamanFormer \cite{koyun_ramanformer_2024} & \nearbest{96.48\% (± 3.12)}  & 77.22\% (± 2.95) & 92.27\% (± 1.68)  & 44.39\% (± 3.38)  & 78.0\% (± 1.02) \\ 
\hline
\end{tabular}%
}
\end{table*}

\begin{table*}[!ht]%
\caption{F1 score of the chosen models on different subsets of the MLROD test dataset. Values are reported as mean $\pm$ standard deviation across five independent final evaluation runs. The F1 score here refers to the macro-averaged F1 score. $\uparrow$ indicates that higher values correspond to better performance for the reported metrics. Boldface highlights the best-performing model in each column. Underlined values indicate models whose mean plus one standard deviation equals or exceeds the highest mean in each column  \label{tab:MLROD_results_f1}}
\resizebox{\textwidth}{!}{%
\begin{tabular}{llllll}
\hline
\textbf{Model} & \textbf{Granite 0\% dusty} $\uparrow$ & \textbf{Granite 50\% dusty} $\uparrow$ & \textbf{Gabbro 0\% dusty} $\uparrow$ & \textbf{Gabbro 50\% dusty} $\uparrow$ &  \textbf{Overall} $\uparrow$  \\
\hline
Random Classifier & 0.0307 (± 0.0019)  & 0.031 (± 0.0012)  & 0.0279 (± 0.0015)  & 0.0246 (± 0.0012)  & 0.0438 (± 0.0009) \\
Majority Class Classifier  & 0.1001 (± 0.0)  & 0.0612 (± 0.0) & 0.0 (± 0.0)  & 0.0 (± 0.0)  & 0.0247 (± 0.0)  \\
Random Forest & 0.4795 (± 0.0193)  & 0.2785 (± 0.0102) & 0.4114 (± 0.0319) & 0.1547 (± 0.0015) & 0.5537 (± 0.0122) \\ 
SVC & 0.396 (± 0.0127) & 0.2519 (± 0.0086)  & 0.3866 (± 0.004)  & 0.1551 (± 0.0006)  & 0.581 (± 0.008) \\
Deep CNN  \cite{liu2017deep} & \textbf{0.7284 (± 0.0248)} & \textbf{0.3779 (± 0.0145)} & 0.6661 (± 0.0301) & \textbf{0.1812 (± 0.0095)} & \textbf{0.7132 (± 0.0202)} \\
SANet \cite{deng2021scale} & 0.5543 (± 0.0621)  & 0.3177 (± 0.0211)  & \textbf{0.715 (± 0.1056)}  & 0.1494 (± 0.0072)  & 0.6383 (± 0.0231)  \\ 
RamanNet \cite{ibtehaz2023ramannet} & 0.5273 (± 0.034) & 0.3239 (± 0.0187) & 0.5508 (± 0.0352) & 0.1629 (± 0.0098)  & 0.6808 (± 0.0193) \\  
Transformer \cite{liu2023classification} & 0.5041 (± 0.061) & 0.2953 (± 0.024) & 0.3763 (± 0.0288)  & 0.1506 (± 0.0115)  & 0.5794 (± 0.0394)  \\ 
RamanFormer \cite{koyun_ramanformer_2024} & 0.5693 (± 0.063)  & 0.3328 (± 0.0284) & 0.4076 (± 0.0486)  & 0.1455 (± 0.0137) & 0.6121 (± 0.0234)  \\ 
\hline
\end{tabular}%
}
\end{table*}

Table \ref{tab:MLROD_results_acc} and Table \ref{tab:MLROD_results_f1} show the classification results for the five Raman-specific DL models and two conventional ML models trained and tested on the MLROD, along with the two baselines. All the models outperform the random classifier and majority class classifier baselines. In addition, the DL models excluding the Transformer outperform the two conventional ML models. 

Another important note is that all the models perform substantially worse on the dusty samples, showcasing the sensitivity of these classifiers to interference. Since dust contaminated spectra are not part of the training dataset, this degradation in performance is due to distribution shift and not overfitting on the training dataset. Therefore, the reported overall test results across all the samples are a reflection of model robustness under domain shift rather than in-distribution classification performance.

The SANet model showcases the best overall accuracy performance, over 7 percentage points higher than the worst DL model (Transformer).  The macro-averaged F1 score being significantly lower than the test accuracy across all the subsets of the MLROD test dataset means that one or more classes are performing considerably worse than the dominant classes.

\subsection{Bacteria-ID}
We report the classification results of the models on two tasks: (1) classifying the isolate (2) classifying the empiric treatment in Table \ref{tab:Bacteria_results}. For both the tasks, the models outperform the random classifier and majority class classifier baselines and the Raman-specific DL models outperform the conventional ML models.

Among the five DL models for the first task, the Transformer performs the worst and classification accuracy of all the other trained DL models are within 2 percentage points of each other. This spread is narrower for the second task, with the test accuracies of the best and worst performing DL models being within 1 percentage point of each other. The comparable macro-averaged F1 score and test accuracy for all the models suggest balanced performance across the classes.

\begin{table*}[!ht]
\caption{Test accuracy and F1 score of the chosen models on the test dataset of the Bacteria-ID dataset. Values are reported as mean $\pm$ standard deviation across five independent final evaluation runs. The F1 score here refers to the macro-averaged F1 score. $\uparrow$ indicates that higher values correspond to better performance for the reported metrics. Boldface highlights the best-performing model in each column. Underlined values indicate models whose mean plus one standard deviation equals or exceeds the highest mean in each column \label{tab:Bacteria_results}}
\begin{center}
\resizebox{0.77\textwidth}{!}{%
\begin{tabular}{lllll}
\hline
&\multicolumn{2}{@{}c}{\textbf{30 isolates}} & \multicolumn{2}{@{}c}{\textbf{8 treatments}} \\\cmidrule{2-3}\cmidrule{4-5} 
\textbf{Model} & \textbf{Accuracy} $\uparrow$  & \textbf{F1 score} $\uparrow$  & \textbf{Accuracy} $\uparrow$  & \textbf{F1 score} $\uparrow$ \\
\hline
Random Classifier & 3.45\% (± 0.4)& 0.0344 (± 0.004) &  12.26\% (± 0.73)& 0.1062 (± 0.0055) \\
Majority Class Classifier & 3.33\% (± 0.0)& 0.0022 (± 0.0) &  26.67\% (± 0.0)& 0.0526 (± 0.0) \\
Random Forest & 58.76\% (± 0.78)& 0.5735 (± 0.0078) & 72.33\% (± 0.51)& 0.5891 (± 0.0073) \\
SVC & 78.11\% (± 0.38)& 0.7704 (± 0.0054) & 94.13\% (± 0.23)& 0.9399 (± 0.0026) \\
Deep CNN \cite{liu2017deep} & \textbf{85.89\% (± 0.12)}& \textbf{0.8565 (± 0.0009)} & \textbf{97.12\% (± 0.15)}& \textbf{0.9748 (± 0.0014)} \\
SANet \cite{deng2021scale} & \nearbest{85.62\% (± 0.6)}& \nearbest{0.8547 (± 0.0063)} & 96.7\% (± 0.2)& 0.9718 (± 0.0012) \\
RamanNet \cite{ibtehaz2023ramannet} & 84.06\% (± 0.76)& 0.8392 (± 0.0082) & 96.47\% (± 0.21)& 0.9672 (± 0.0028) \\
Transformer \cite{liu2023classification} & 83.27\% (± 0.62)& 0.827 (± 0.0069) & 96.43\% (± 0.13)& 0.9667 (± 0.0013) \\
RamanFormer \cite{koyun_ramanformer_2024} & 84.63\% (± 0.58)& 0.8424 (± 0.0063) & \nearbest{97.05\% (± 0.14)}& \nearbest{0.9725 (± 0.0025)} \\
\hline
\end{tabular}
}
\end{center}
\end{table*}

\subsection{API Dataset}
We show the results of all the models and baselines on the 32 category classification task in the Dataset Label Space and the 31 category classification task in the Chemical Identity Space of the API dataset in Table \ref{tab:Pharma}. The five DL models and the two conventional ML models outperform the random classifier and the majority class classifier. However, unlike on the other datasets, the ML models perform comparably to the Raman-specific DL models across both the tasks. 

For the former task, the Transformer is the worst performing model and the accuracy of the Deep CNN, RamanNet, RamanFormer and SVC are all within a percentage point of each other followed by the Random Forest and SANet being 1 and 2 percentage points clear from this pack respectively. The macro-averaged F1 score being similar to the test accuracy is indicative of relatively uniform classification performance across the different categories for all the 5 models.  

\begin{figure*}[ht!]
\centering
\includegraphics[width=\textwidth]{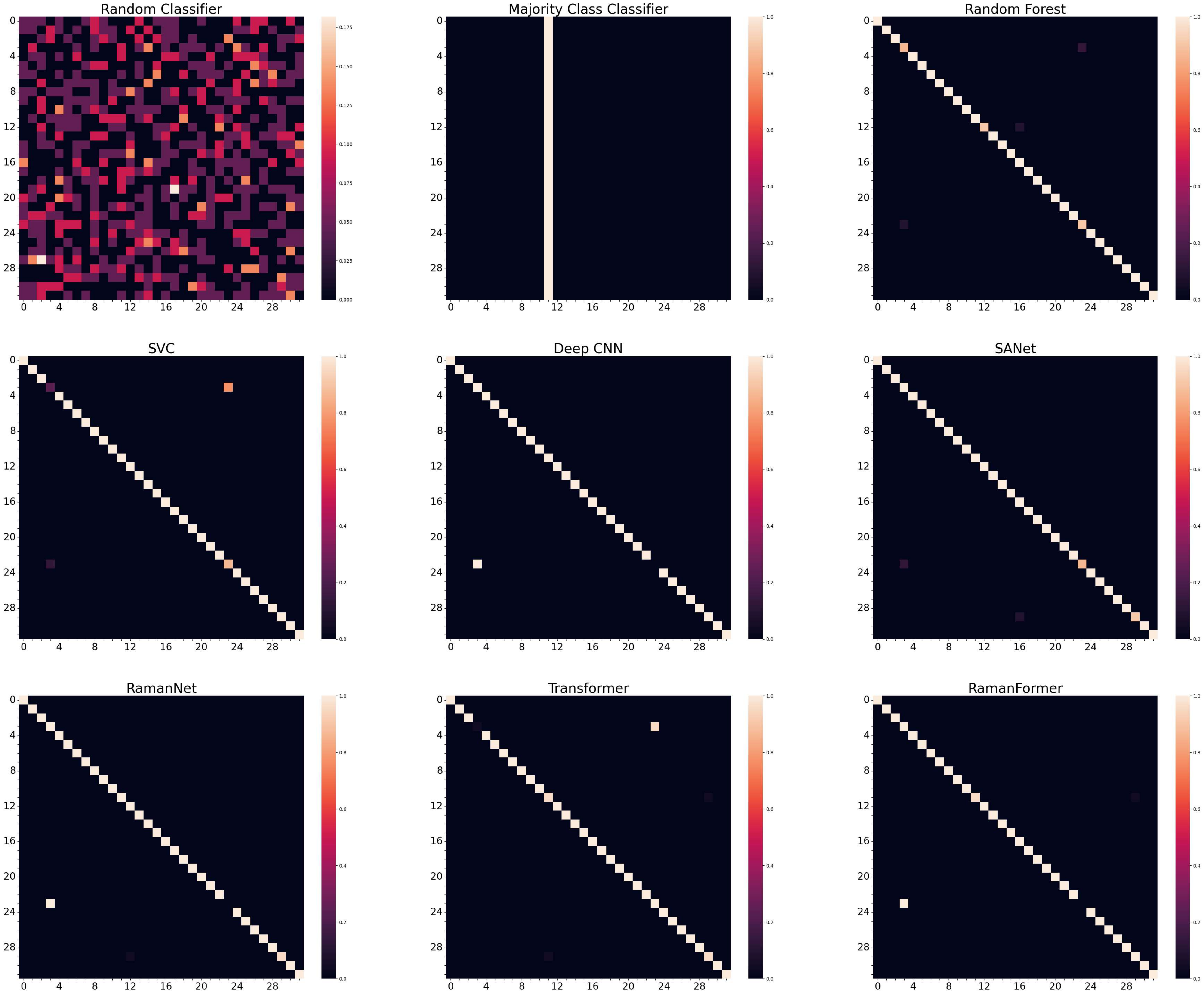}
\caption{Class-normalized confusion matrices for the 32 category classification task using the API test set on Dataset Label Space. Rows indicate true class labels and columns indicate predicted class labels. The matrices compare the random classifier, majority class classifier and the benchmarked machine learning and deep learning models. Brighter diagonal entries correspond to a higher classification accuracy. Supplementary Table 7 lists the materials associated with each class label. Most of the misclassifications for all the models are between class labels 3 and 23 corresponding to 4-Methyl-2-pentanone and Methyl Isobutyl Ketone respectively, which are chemically identical. 
\label{fig4}}
\end{figure*}

We observe a consistent increase in performance for the 31 category evaluation task, with all the models achieving a test accuracy greater than 99\%. This improvement across the models suggests that most of the errors in the 32-class setting arise from confusion between synonymous labels rather than a failure to learn chemically meaningful representations. This observation is further strengthened by the confusion matrix in Figure \ref{fig4}, which shows most of the misclassifications occurring between the Raman spectra of the chemically identical compounds 4-Methyl-2-pentanone and Methyl Isobutyl Ketone.

\begin{table*}[!ht]
\caption{Test accuracy and F1 score of the chosen models on the test dataset of the API dataset. Values are reported as mean $\pm$ standard deviation across five independent final evaluation runs. The F1 score here refers to the macro-averaged F1 score. $\uparrow$ indicates that higher values correspond to better performance for the reported metrics. Boldface highlights the best-performing model in each column. Underlined values indicate models whose mean plus one standard deviation equals or exceeds the highest mean in each column \label{tab:Pharma}}
\begin{center}
\resizebox{0.77\textwidth}{!}{%
\begin{tabular}{lllll}
\toprule
&\multicolumn{2}{@{}c}{\textbf{Dataset Label Space}} & \multicolumn{2}{@{}c}{\textbf{Chemical Identity Space}} \\\cmidrule{2-3}\cmidrule{4-5} \\
\textbf{Model} & \textbf{Accuracy} $\uparrow$  & \textbf{F1 score} $\uparrow$  & \textbf{Accuracy} $\uparrow$  & \textbf{F1 score} $\uparrow$ \\
\midrule
Random Classifier & 3.32\% (± 0.73)& 0.0332 (± 0.0073) & 3.32\% (± 0.79)& 0.0326 (± 0.0078) \\
Majority Class Classifier & 3.12\% (± 0.0)& 0.0019 (± 0.0) & 6.25\% (± 0.0)& 0.0038 (± 0.0) \\
Random Forest & 98.66\% (± 0.17)& 0.9867 (± 0.0017) & 99.35\% (± 0.21)& 0.9936 (± 0.002) \\
SVC & 97.16\% (± 0.16)& 0.968 (± 0.0016) & \textbf{100.0\% (± 0.0)}& \textbf{1.0 (± 0.0)} \\
Deep CNN \cite{liu2017deep} & 97.02\% (± 0.16)& 0.965 (± 0.0057) & \textbf{100.0\% (± 0.0)}& \textbf{1.0 (± 0.0)} \\
SANet \cite{deng2021scale} & \textbf{99.63\% (± 0.21)}& \textbf{0.9963 (± 0.0021)} & 99.86\% (± 0.13)& 0.9985 (± 0.0013) \\
RamanNet \cite{ibtehaz2023ramannet} & 96.73\% (± 0.2)& 0.9593 (± 0.0049) & \nearbest{99.83\% (± 0.17)}& 0.9983 (± 0.0016) \\
Transformer \cite{liu2023classification} & 96.14\% (± 0.36)& 0.9566 (± 0.0029) & 99.29\% (± 0.24)& 0.9927 (± 0.0024) \\
RamanFormer \cite{koyun_ramanformer_2024} & 96.85\% (± 0.11)& 0.9628 (± 0.0048) & 99.91\% (± 0.07)& 0.9991 (± 0.0007) \\
\bottomrule
\end{tabular}
}
\end{center}
\end{table*}

\section{Discussions} \label{sec7}
Benchmarking plays an important role in identifying the strengths and weaknesses of different approaches for a given problem. In this work, we compare the performance of five supervised deep learning Raman models and two conventional ML models across three open-source Raman spectroscopy datasets under consistent training configurations. With this setup, SANet and Deep CNN demonstrate the strongest performance across the datasets. In contrast, the transformer-based models underperform relative to the strongest CNN-based models under the unified training protocol used in this benchmark. This result should be interpreted within the limits of the present experimental design as transformer performance may depend on training choices that were not exhaustively explored here such as learning-rate warmup, weight decay, dropout tuning and longer training schedules. The stronger performance of RamanFormer (4M parameters) compared with the larger ViT-style Transformer (85M parameters) suggests that model scale and Raman-specific architectural choices can also play an important role. 

Another note is that the conventional ML models underperform the DL models on the larger and more diverse MLROD and Bacteria-ID datasets. This suggests that the Raman-specific DL architectures are better able to capture discriminative spectral patterns in these more challenging settings. However, on the smaller API dataset, where the training and test sets are drawn from the same source distribution, the conventional ML models perform comparably to the DL models. This indicates that simpler ML models can remain competitive when the dataset is smaller and the train-test distribution shift is limited.

As shown in Table \ref{tab:MLROD_results_acc} and Table \ref{tab:MLROD_results_f1}, variations in the spectroscopic hardware and acquisition set-up lead to covariate shifts in the Raman spectra captured in the train and test datasets. Calibration Transfer and Maintenance (CTM) methods in Chemometrics \cite{nikzad2021chemometrician} can achieve adaptation between the source(train) and target(test) domains. Many of these CTM methods are not viable for the classification task in MLROD as they require labels for the samples in the target(test) domain \cite{lai2025calibration}  \cite{boucher2017proximal}. Recent works have proposed unsupervised domain adaptation frameworks \cite{umprecht2025unsupervised} \cite{zhang2025unsupervised} but their effectiveness beyond the originally reported datasets are yet to be independently evaluated. 

Another approach to solving this issue is by using self-supervised frameworks on large amounts of unlabeled data to learn more robust representations of Raman spectra.  SMAE \cite{ren2025self} employs a transformer-based encoder and decoder architecture where the self-supervised pretraining task involves recovering the original spectrum from a randomly masked spectrum. The effectiveness of the learned representations was showcased by achieving an unsupervised clustering accuracy of 80.56\% for the 30 class pathogenic species identification problem, where K-means clustering was applied to the extracted features of the Raman spectra in the test set of the Bacteria-ID dataset. SemiRaman \cite{sun2025semiraman} is another framework that combines self-supervised contrastive learning with semi-supervised learning. It involves pretraining an encoder with several contrastive loss functions on augmented views of the same spectrum. The pretrained encoder is then fine-tuned in a multi-stage manner using limited labeled spectra and pseudo-labels. The representations learned in this semi-supervised framework show strong classification performance while using just 5\% of the labeled data of a subset of the Bacteria-ID dataset.

However, these approaches primarily address the challenge of limited labeled Raman spectral data and do not explicitly take into account the variations arising from differences in instrumentation and experimental setups. Such variability can lead to significant distribution shifts, which can cause degradation of model performance during testing. Foundation models are designed to be quite robust to input distribution shifts by learning representations from large and diverse datasets. The Deep-spectral Component Filtering (DSCF) \cite{xue2025deep} is a foundation model that was developed through spectral component resolvable learning on over a million simulated and experimental spectra spanning several experiments and spectroscopic modalities (UV, IR and Raman). The authors showcased strong performance of DSCF in multi-label classification with concentration estimation using simulated and experimental SERS spectra. However, its performance has not yet been evaluated on standard open-source Raman spectroscopy datasets.

The development of DSCF highlights the need for large-scale and diverse training data. Existing open-source Raman datasets are often restricted in size, chemical diversity or experimental variability. Creating large, curated experimental Raman spectral datasets that span multiple instruments, materials and measurement settings is key to developing a Raman-specific foundation model. These models could serve as a reusable backbone for a wide range of downstream tasks through lightweight fine-tuning using relatively small labeled datasets. This would significantly reduce annotation and training costs as the models would no longer need to be trained from scratch.

\section{Conclusions}
In this work, we presented a benchmark of five supervised deep learning models for Raman spectra classification and two conventional ML models evaluated across three public datasets. These datasets capture different real-world challenges like domain shift due to acquisition variability (MLROD), multi-task clinical labeling (Bacteria-ID) and high-accuracy multi-category classification (API). All the models were evaluated using a unified experimental protocol to ensure fair comparison.

In MLROD, we observed a significant degradation in performance for all the models on the 50\% dusty test samples compared to 0\% dusty test samples, implying brittleness to background interference and acquisition shift. In Bacteria-ID, isolate classification accuracy sits between 83-86\%, while antibiotic treatment prediction is 96-98\% for the five DL models with the ML models performing a step below them. Finally for the API dataset, all the models achieved near-ceiling accuracies between 99-100\%. Overall, SANet and Deep CNN demonstrate the best overall performance across the datasets, with the other Raman-specific deep learning models not too far behind, followed by the two conventional machine learning models under this unified experimental protocol.

A key trend is that simpler machine learning methods can remain competitive when the classification task has limited train-test distribution shift, while Raman-specific deep learning architectures offer advantages in larger datasets with train and test sets not drawn from the same source.

The results of this benchmarking experiment have also shown that classifying test samples that are in-distribution to the training dataset is significantly easier than test samples suffering from distribution shift due to changes in instruments and acquisition conditions, and additional contaminants. While this study benchmarks only five Raman-specific supervised DL models and two conventional ML models while relying on minimal spectral pre-processing, it establishes a transparent and reproducible baseline for evaluating supervised Raman spectra classifiers. We hope this benchmark will facilitate more rigorous comparisons, thereby enabling researchers to identify effective design choices and develop improved models in the future.

\section*{Author contributions}

Adithya Sineesh conducted the Literature Survey, performed the benchmarking experiments and wrote the Benchmark Models, Results and Discussions Section.

Akshita Kamsali stylistically edited the Literature Survey along with authoring the Abstract, Introduction, Datasets, Methodology and Conclusion sections.

\section*{Data Availability}
The links to the Raman datasets, implementation of the model architectures, training scripts and evaluation code can be found in the GitHub repo – \url{https://github.com/asineesh/Benchmark_Raman_DeepLearning}. The trained models and log files are available at \url{https://doi.org/10.5281/zenodo.19701494}.\\

Supplementary information (SI) contains the hyperparameter grids used by the conventional ML models, training wall-clock times
for all the evaluated models, class-wise distributions for the MLROD, Bacteria-ID, and API datasets, and additional class-normalized confusion matrices for each dataset/task and model pair. See DOI: \url{https://doi.org/10.1039/d6dd00044d}.

\section*{Acknowledgments}
The authors would like to thank Dr. Avinash Kak for valuable discussions, guidance and access to computational resources that supported this work. The authors would also like to thank Dr. Rahul Deshmukh for his feedback on the manuscript.

\section*{Financial disclosure}

None reported.

\section*{Conflict of interest}

The authors declare no potential conflict of interests.

\bibliographystyle{abbrv}
\bibliography{references}

\end{document}